\documentclass[runningheads]{llncs}
\usepackage{graphicx}

\usepackage{amsfonts}
\usepackage{amsmath}
\usepackage{booktabs}
\usepackage{multirow}
\usepackage{soul}
\usepackage{subcaption}
\usepackage{subfiles}

\newcommand\doubleplus{+\kern-1.3ex+\kern0.8ex}

\begin{document}
\title{Message Passing Neural Networks \\ for Traffic Forecasting}
%
\titlerunning{MPNN for Traffic Forecasting}
%
\author{
Arian Prabowo\inst{1,2} \and
Hao Xue\inst{3} \and
Wei Shao\inst{4} \and
Piotr Koniusz\inst{2,5} \and
Flora D. Salim\inst{3}
}

\authorrunning{A. Prabowo et al.}
\institute{
RMIT University, Australia \and
Data61/CSIRO, Australia \and
UNSW, Australia \and
UC Davis, USA \and
ANU, Australia
}
%
\maketitle              
\begin{abstract}
A road network, in the context of traffic forecasting, is typically modeled as a graph where the nodes are sensors that measure traffic metrics (such as speed) at that location.
Traffic forecasting is interesting because it is complex as the future speed of a road is dependent on a number of different factors.
Therefore, to properly forecast traffic, we need a model that is capable of capturing all these different factors.
A factor that is missing from the existing works is the node interactions factor.
Existing works fail to capture the inter-node interactions because none are using the message-passing flavor of GNN, which is the one best suited to capture the node interactions
This paper presents a plausible scenario in road traffic where node interactions are important and argued that the most appropriate GNN flavor to capture node interactions is message-passing.
Results from real-world data show the superiority of the message-passing flavor for traffic forecasting.
An additional experiment using synthetic data shows that the message-passing flavor can capture inter-node interaction better than other flavors.
\end{abstract}
\section{Introduction}

Traffic forecasting is the task of predicting future traffic measurements (e.g. volume, speed, etc.) in a road network based on historical data.
It is an interesting task from the perspective of both the broader societal impact of its application, as well as from a theoretical perspective.
As a part of intelligent transport systems, better traffic forecasting can lead to more efficient designs and management of transportation networks.
The downstream impacts include decreases in energy usage and reductions of the negative economic and psychological impacts of congestion.

Recently, traffic forecasting also gathers the interests of the research community \cite{jiang2021DL-TrAFF}.
This is because, from the theoretical perspective, traffic forecasting is interesting because it is complex as the future speed of a road is dependent on a number of different factors.
Therefore, to properly forecast traffic, we need a model that is capable of capturing all these different factors.
An example of one such factor is each node's unique periodical pattern.
To capture this behavior, AGCRN \cite{bai2020AGCRN} learned a unique set of weights for each node.

Another factor is the past speed of neighboring nodes.
To capture information from the neighboring nodes, many models used Graph Neural Networks (GNNs) which come in three flavors: convolutional, attentional, and message-passing \cite{bronstein2021geometric}.
For example, DCRNN\cite{li2018DCRNN}, STGCN\cite{yu2018STGCN}, Graph WaveNet\cite{wu2019GraphWaveNet}, MTGNN\cite{wu2020MTGNN}, AGCRN\cite{bai2020AGCRN} used different variations of convolutional GNNs, while ASTGCN\cite{guo2019ASTGCN} and GMAN\cite{zheng2020gman} used different variations of attentional GNNs.

However, in this paper, we argue that there is another factor that is missing from the existing works, and that is the node interactions factor.
Existing works fail to capture the inter-node interactions because none are using the message-passing flavor of GNN, which is the one best suited to capture node interactions.

Our contributions are as follows:
\begin{itemize}
    \item This paper presents a plausible scenario in road traffic where nodes interactions are important.
    \item We then argued that the most appropriate GNN flavor to capture node interactions is message-passing.
    \item We perform several experiments using real-world data to show the superiority of the message-passing flavor for traffic forecasting.
    \item We perform an additional experiment using synthetic data to show that the message-passing flavor can capture inter-node interaction better than other flavors.
\end{itemize}

In the next section, we are going to formally describe the traffic forecasting task, node interactions in the context of traffic, and GNN flavors in the context of node interactions.
Then, we briefly review the existing works on traffic forecasting, from the perspective of GNN flavors.
Next, we describe the models we use in our experiments.
These are: Graph WaveNet \cite{wu2019GraphWaveNet}, the backbone architecture for our experiment;
Diffusion convolution \cite{li2018DCRNN} a part of Graph WaveNet representing the convolutional flavor;
Graph Attention Network (GAT) \cite{velickovic2018GAT} representing the attentional flavor; and
Message Passing Neural Network (MPNN) \cite{gilmer2017MPNN} representing the message-passing flavor.
We then present and describe the results of two sets of experiments;
one set on real-world data and another on synthetic data.
Finally, we discuss the ethical implication of our work and present our conclusion.

\section{Preliminary}
\subsection{Traffic Forecasting Problem Definition}
\label{sec:def}

A traffic dataset is defined as a tensor
$\mathbf{\chi} \in \mathbb{R}^{D \times N \times L}$
where
$D$ is the value of different traffic metrics (such as speed, flow, and density) at a particular node,
$N$ is the number of nodes, and
$L$ is the number of timesteps.
The dataset can be accompanied with adjacency matrices $A \in \mathbb{R}^{B,N,N}$,
where
$B$ is the number of different adjacency matrices.
Since the edges can be weighted (usually as the inverse of distance) the adjacency matrix contains real values instead of binary.
Moreover, because there are more than one way to construct adjacency matrices, some works use multiple adjacency matrices.

The traffic forecasting task is to perform multi-step forecast of the near future traffic metrics $\mathbf{\chi}[:,:,l+L_{FH}:l+L_{FH}+L_{FW}]$
based on the recent past traffic metrics
$\mathbf{\chi}[:,:,l-L_{OW}:l]$
and the adjacency matrix
$\mathbf{A}$.
The typical value is 12 timesteps (1 hour) for observation and forecasting window.
This is formalized as follows:
\begin{align}
\mathbf{\chi}[:,:,l+L_{FH}:l+L_{FH}+L_{FW}]
= f(
\mathbf{\chi}[:,:,l-L_{OW}:l]
, \mathbf{A})
\end{align}
where
$l$ is the forecasting time,
$L_{OW}$ is the observation window,
$L_{FH}$ is the forecasting horizon,
and
$L_{FW}$ is the forecasting window.
Figure \ref{fig:problem_def} in the Supplementary Material shows the problem definition visually.

\subsection{Node interactions}


In this subsection, we are going to present a simplistic but illustrative scenario where node interaction is an important factor in road traffic.
Consider a subgraph consisting of two nodes, $u$ and $v$, and a directed edge $(u,v)$, representing a one way street.
Each node has a corresponding binary traffic state $x^t_u$ and $x^t_v$ at time $t$ with the options being: congested or free-flow.
The binary function $x^{t+1}_v(x^t_u,x^t_v)$ represents the future traffic at node $v$ at time $t+1$.
Generally, we can decompose this function into three different terms:
$
x^{t+1}_v(x^t_u,x^t_v)
=
f(x^t_u)+
g(x^t_v)+
h(x^t_u,x^t_v)
$
where
$f(x^i_u)$ is the term representing the influence of the neighboring nodes (in this case, there is only one),
$g(x^i_v)$ is the term representing the influence of node $v$ onto itself, and
$h(x^i_u,x^i_v)$ is the term representing the node interactions.
(Note that the idea is that $h(\cdot)$ is whatever is left from $f(\cdot)$ and $g(\cdot)$.)

Now, let's imagine a situation where the timestep is relatively short compared to the clearing of congestion at a node, but sufficiently long for congestion from node $u$ to be able to reach node $v$ in a timestep.
Then the future of node $v$ can be modeled with an AND function:
$
x^{t+1}_v(x^t_u,x^t_v)
=x^t_u \wedge x^t_v
$.
This behavior cannot be reduced in terms of combinations of $f(\cdot)$ and $g(\cdot)$ alone.
Thus, we show that it is plausible that a scenario in traffic exists where the node interaction $h(\cdot)$ is important.



\subsection{GNN flavors}

\begin{figure*}[htb]
    \centering
    \includegraphics[width=\textwidth]{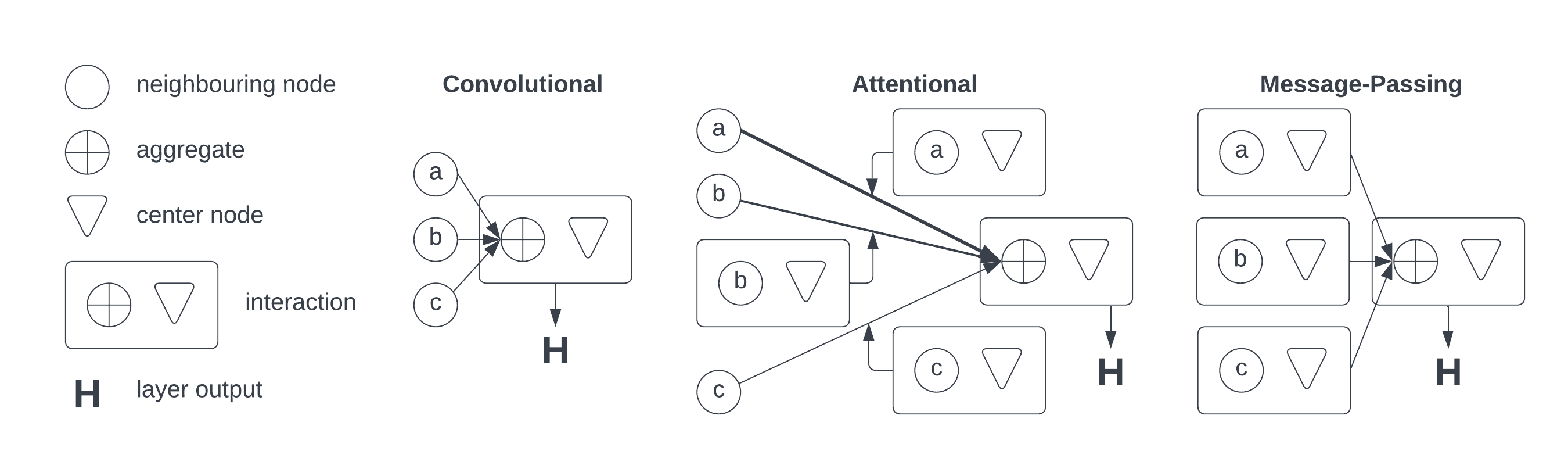}
    \caption{
    Visualization of the three flavors of GNN \cite{bronstein2021geometric}.
    The convolutional flavor lacks direct interactions between the center node and its neighbors.
    The attentional flavor uses the center and neighbor nodes interactions to dynamically adjust the attention weights of the neighboring nodes.
    Only the message-passing flavor allows direct interactions between the center and the neighboring nodes.
    }
    \label{fig:flavors}
\end{figure*}

One popular taxonomy of GNNs is the three flavors\cite{bronstein2021geometric}: convolutional, attentional, and message-passing, visualized in Figure \ref{fig:flavors}.
The idea behind all GNN layers is to update a node's latent features based on its neighborhood (including itself).
They all have the general form of:
\begin{align}
\mathbf{h}^{i+1}_u=
\boldsymbol\phi(
\mathbf{h}^{i}_u,
\mathcal{H}^{i}_u
\left( \left\{
\mathbf{h}^{i}_v|v\in u \cup \mathcal{N}_u
\right\} \right)
)
\end{align}
where
$\boldsymbol\phi(\cdot)$ is a learnable non-linear transformation such as Multi-Layer Perceptron (MLP),
$\mathbf{h}^{i}_u$ is the latent feature of node $u$ at layer $i^\mathrm{th}$, 
$\mathcal{H}^{i}_u(\cdot)$ is a function that acts on the neighbourhood, and
$\mathcal{N}_u$ is a set containing all the neighbours of node $u$.
The key difference between the flavors is in their choice of $\mathcal{H}^{i}_u(\cdot)$ that decides what the central node $u$ is allowed to interact with.

In a convolutional GNN layer,
\begin{align}
\mathcal{H}^{i}_u=
\bigoplus_{\mathcal{N}_u} c_{uv} \boldsymbol\psi(\mathbf{h}^{i}_v)
\end{align}
where
$\boldsymbol\psi(\cdot)$ are learnable non-linear transformations such as MLP,
$\bigoplus$ is a permutation-invariant aggregator such as mean-pooling, max-pooling, or summation, and
$c_{uv}$ is a learnable constant.
Here, the central node $u$ is only allowed to interact with the aggregate of the neighborhood.
There are no node interactions.

In an attentional GNN layer,
\begin{align}
\mathcal{H}^{i}_u=
\bigoplus_{\mathcal{N}_u} \alpha(\mathbf{h}^{i}_u,\mathbf{h}^{i}_v) \boldsymbol\psi(\mathbf{h}^{i}_v)
\end{align}
where
$\alpha(\cdot,\cdot)$ is a scalar function representing attention mechanism \cite{velickovic2018GAT}.
It dynamically modulates the contribution of each of the node neighbours.
Since the scalar attention is calculated from the node interactions $a(\mathbf{x}_i,\mathbf{x}_j)$, when combined with the multi-head technique, the linear combinations of the different heads can approximate the full inter-nodes interactions.

In a message-passing GNN layer,
\begin{align}
\mathcal{H}^{i}_u=
\bigoplus_{\mathcal{N}_u} \boldsymbol\psi(\mathbf{h}^{i}_u \doubleplus \mathbf{h}^{i}_v)
\end{align}
where
$\doubleplus$ is the binary concatenation operator.
Here, $\mathcal{H}^{i}_u(\cdot)$ captures the node interactions before the aggregations.
This ensures that all the important information arising from the interactions between pairs of nodes is preserved.
Thus, in tasks where it is important to capture interactions between nodes, such as traffic, message-passing is the most appropriate flavor.

\section{Related Works}

\begin{table}[ht]
\centering
\caption{GNN flavors of existing works}
\label{tab:rw}
\setlength{\tabcolsep}{1em}
\begin{tabular}{@{}llccc@{}}
\toprule
 & Model         & Conv. & Att. & M-P  \\ \midrule
IJCAI 2018 & STGCN\cite{yu2018STGCN}         &  \checkmark   &             &                        \\
ICLR 2018 & DCRNN\cite{li2018DCRNN}         & \checkmark              &             &                       \\
IJCAI 2019 & Graph WaveNet\cite{wu2019GraphWaveNet} &  \checkmark             &             &                        \\
AAAI 2019 & ASTGCN\cite{guo2019ASTGCN}        &               & \checkmark            &                        \\
AAAI 2020 & GMAN\cite{zheng2020gman}          &               &  \checkmark           &                        \\
SIGKDD 2020 & MTGNN\cite{wu2020MTGNN}         &  \checkmark             &             &                        \\
NeurIPS 2020& AGCRN\cite{bai2020AGCRN}         &  \checkmark             &             &                       \\
PAKDD 2021& AGCAN\cite{li2021adaptive}         &            & \checkmark               &                       \\
PAKDD 2022& T\textsuperscript{2}-GNN\cite{shen2022two}         &  \checkmark             &             &                       \\
 & \textbf{Ours}          &               &             & \checkmark                       \\ \bottomrule
\end{tabular}
\end{table}

Road traffic has spatial and temporal elements.
However, in this section, we are only going to focus on how the models proposed in the existing works propagate spatial information, and the GNN flavors used.
This information is summarized in Table \ref{tab:rw}.
For a more thorough review of this topic, check the survey \cite{tedjopurnomo2020survey} and the benchmark paper \cite{jiang2021DL-TrAFF}.
Earlier work such as LSTNet \cite{lai2018LSTNet} did not model any spatial dependencies.

\subsection{Convolutional}

STGCN \cite{yu2018STGCN} focused on making the multi-hop spatial information propagation efficient.
To that end, they choose to use the Chebychev polynomial to approximate the spectral convolution kernels \cite{defferrard2016ChebNet}.
At about the same time, DCRNN \cite{li2018DCRNN} argued that a weighted and directed graph is a better model of the traffic network.
To that end, they formulated a spatial convolution operation, called diffusion convolution, that works on weighted and directed graphs.
They also showed that, when a graph is unweighted and non-directed, the diffusion convolution becomes equivalent the spectral convolution.

Graph WaveNet \cite{wu2019GraphWaveNet} also used the diffusion convolution proposed by in DCRNN \cite{li2018DCRNN}.
However, it argued that there are hidden spatial dependencies that are not captured by the adjacency matrix constructed from the physical road network.
Instead, it proposed the construction of self-adjacency matrices that learn the hidden spatial structure from data.
In the absence of information about the spatial connectivity, these self-adjacency matrices can be used alone.

MTGNN \cite{wu2020MTGNN} extended this idea of learning the graph structure from the data, to any multivariate time series instead of just traffic.
Regarding the propagation of spatial information, they used a multi-hop spatial Graph Convolutional Network (GCN) \cite{klicpera2018predict}.
Furthermore, they argued that it is important to dynamically adjust the contributions of neighborhoods with different numbers of hops.
This was done by using a gating mechanism.

AGCRN \cite{bai2020AGCRN} argued that every node has a unique behavior, and thus, there should be a different set of weights for every node.
To prevent over-fitting and the explosion in the number of parameters, they proposed matrix factorization of the GCN parameters.
T\textsuperscript{2}-GCN \cite{shen2022two} decomposed the dynamic spatial correlations in a traffic network into seasonal static and acyclic dynamic components.
Because the two components have different dynamic spatial correlations, they used two parallel GCN towers, each containing a dynamic graph generation layer.

\subsection{Attentional}

ASTGCN \cite{guo2019ASTGCN} pointed out the lack of attentional flavor in the literature then.
Spatially, they proposed an attention mask over each of the Chebyshev terms.
GMAN \cite{zheng2020gman} argued that the inter-node correlations changes over the typical forecasting window (1 hour).
Thus, to capture these fast changes, they used multi-head attention \cite{Vaswani2017transformer} to make the graph convolution steps dynamic.
AGCAN \cite{li2021adaptive} observed that adjacent nodes do not always have similar behavior due to differences in road attributes (like road width).
On the other hand, non-adjacent nodes could have very similar traffic patterns, also due to road attribute similarities (like point-of-interest).
To this end, they proposed a hierarchical spatial attention module.

\subsection{Message-passing}
To the best of our knowledge, we are the first to try the message-passing flavor to the traffic forecasting task, as shown in Table \ref{tab:rw}.
MPNN \cite{gilmer2017MPNN} was the first to be introduced as a generalization of several different GNN.
The first implementation was in the domain of quantum chemistry and remained as a popular application up to recent years.
For example, DimeNet \cite{Gasteiger2020Directional} added a way for the model to access information regarding the angle between bonds in a molecule.

\section{Methods}

\begin{figure*}[htb]
    \centering
    \includegraphics[width=\textwidth]{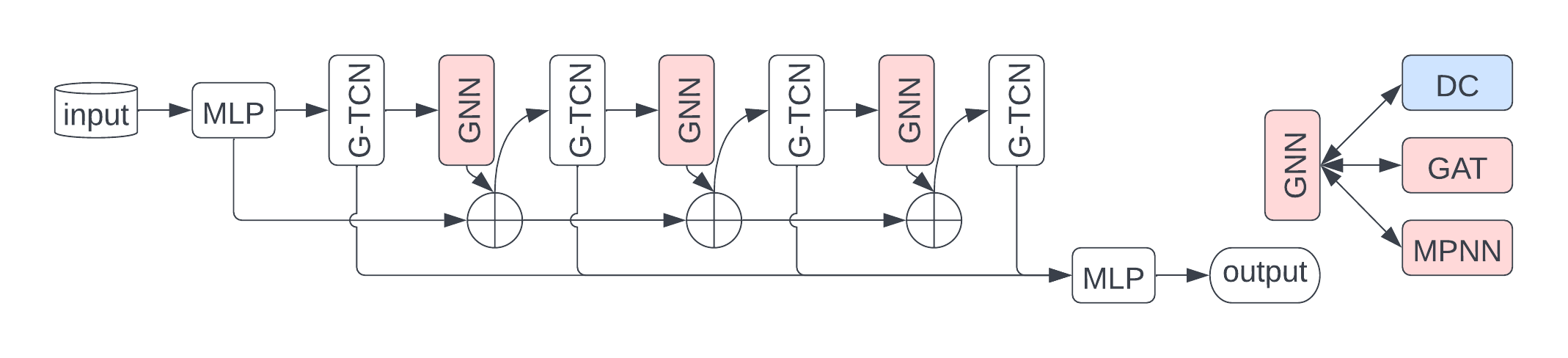}
    \caption{
    Simplified architecture diagram of Graph WaveNet backbone.
    Number of layers is only illustrative.
    G-TCN is a gated temporal convolutional network layer \cite{van2016wavenet}.
    $\bigoplus$ is a summation operation.
    In the original paper \cite{wu2019GraphWaveNet}, the GNN of choice is diffusion convolution (DC) \cite{li2018DCRNN}.
    In this paper, we replace the DC with GAT \cite{velickovic2018GAT} or MPNN \cite{gilmer2017MPNN}.
    }
    \label{fig:gwn_arch}
\end{figure*}

To evaluate the impact of different flavors of GNN for traffic forecasting, we first pick the current state-of-the-art architecture as the backbone.
Based on a recent benchmark \cite{jiang2021DL-TrAFF}, we picked Graph WaveNet \cite{wu2019GraphWaveNet} as the backbone as it remains state-of-the-art.
Its usage as a backbone remained popular in recent works \cite{shen2022two}.
Next, we replaced the GNN module of Graph WaveNet with other flavors of GNN as shown in Figure \ref{fig:gwn_arch}.
In particular, we picked Graph Attention Network (GAT) \cite{velickovic2018GAT} and Message Passing (MPNN) \cite{gilmer2017MPNN} as representatives of the attentional and message-passing flavors respectively, to replace the diffusion convolution \cite{li2018DCRNN} in Graph WaveNet.
This way, we can attribute any changes in performance to the different flavors alone.

For the rest of this section, we describe diffusion convolution, GAT, and MPNN.
A short summary of our Graph WaveNet backbone can be found in the Supplementary Material \ref{apx:gwn}.

\subsection{Diffusion convolution with self-adaptive adjacency matrix} 
DCRNN \cite{li2018DCRNN} introduced diffusion convolution for directed graphs.
Graph WaveNet \cite{wu2019GraphWaveNet} added a self-adaptive adjacency matrix term into the diffusion convolution.
The combined version is as follows:
\begin{align}
\mathbf{h}^{[l+1]} =
\mathbf{B} +
\sum^K_{k=0} \mathbf{P}^k_f \mathbf{h}^{[l]} \mathbf{W}_{k,1}
+ \mathbf{P}^k_b \mathbf{h}^{[l]} \mathbf{W}_{k,2}
+ \Tilde{\mathbf{A}}^k \mathbf{h}^{[l]} \mathbf{W}_{k,3}
\end{align}
\cite{li2018DCRNN}
where
$\mathbf{h}^{[l]}$ and $\mathbf{h}^{[l+1]}$ are the input and output of the $l^\mathrm{th}$ module,
$\mathbf{P}^k_f$ and $\mathbf{P}^k_b$ are the forward and backward normalized adjacency matrices respectively, to the power of $k$,
$\mathbf{W}_{k,1}$, $\mathbf{W}_{k,2}$, $\mathbf{W}_{k,3}$, and $\mathbf{B}$ are the learnable parameters,
$K$ is the number of hops, and
$\Tilde{\mathbf{A}}$ is the self-adaptive adjacency matrix.
The adjacency matrices are normalized as follows:
$
\mathbf{P}_f =
\mathbf{A} / \mathrm{rowsum}(\mathbf{A})
$ and
$
\mathbf{P}_b =
\mathbf{A}^T / \mathrm{rowsum}(\mathbf{A}^T)
$.
The self-adaptive adjacency matrix is defined as follows:
$
\Tilde{\mathbf{A}} =
\mathrm{SoftMax}(
\mathrm{ReLU}(
\mathbf{E}_1 \mathbf{E}_2^T
))$
where
$
\mathbf{E}_1,\mathbf{E}_2 \in
\mathbb{R}^{N \times c}
$
are learnable parameters,
$N$ is the number of nodes, and
$c$ is the size of latent feature.

\subsection{Graph Attention Network (GAT)}
Based on the success of attention \cite{bahdanau2015attention} and multi-headed self-attention mechanism \cite{Vaswani2017transformer}, GAT \cite{velickovic2018GAT} extended this to a graph structured data.
We picked GAT as the representative of the attentional flavor.
We used this PyTorch implementation: \url{github.com/Diego999/pyGAT}.

The forward function of GAT is:
\begin{align}
\Tilde{\mathbf{h}}^{[l]}_i =
\underset{m=1}{\overset{M}{\doubleplus}}
\mathrm{ELU} \left(
\sum_{j\in \mathcal{N}_i} \alpha_{m,i,j}
\mathbf{W}_{1,m}
\mathbf{h}^{[l]}_j
\right)
\end{align}
where
$\Tilde{\mathbf{h}}^{[l]}_i$ is the output of the GAT layer for the center node $i$,
$\doubleplus$ represents concatenation of $M$ different heads,
$\mathrm{ELU}(\cdot)$ is the exponential linear unit \cite{Clevert2016FastAA} acting as the activation function,
$\mathcal{N}_i$ is the set containing all the neighbours of node $i$,
$\alpha_{m,i,j}$ is the attention coefficient of head $m$ from node $i$ to node $j$,
$\mathbf{W}_{1,m}$ is the learnable weights, and
$\mathbf{h}^{[l]}_j$ is the latent feature of node $j$.
The attention coefficients are calculated as follows:
\begin{align}
\alpha_{i,j} =
\mathrm{SoftMax} \left(
\mathrm{LeakyReLU} \left(
\left[ \mathbf{W}_2 \mathbf{h}^{[l]}_i \doubleplus \mathbf{W}_2 \mathbf{h}^{[l]}_j \right] \mathbf{W}_3
\right) \right)
\end{align}
where 
$\mathbf{W}_2$ and $\mathbf{W}_2$ are also learnable weights.

Due to the concatenation operation of the multiple heads, $\Tilde{\mathbf{h}}^{[l]}_i$ had a different size than the input.
To rectify this, we add another MLP to reduce the size back.
$\mathbf{h}^{[l+1]}_i =
\mathrm{MLP}(\Tilde{\mathbf{h}}^{[l]}_i)$

\subsection{Message-Passing Neural Network (MPNN)}
We chose a simple message-passing layer and we make our own implementation.
It is defined as follows:
\begin{align}
\mathbf{h}^{[l+1]}_i =
\mathrm{MLP}_1\left(
\sum_{j \in \mathcal{N}_i}
\mathrm{MLP}_2\left(
\mathbf{h}^{[l]}_i
\doubleplus \mathbf{h}^{[l]}_j
\doubleplus P_{f,i,j}
\doubleplus P_{b,i,j}
\doubleplus \Tilde{A}_{i,j}
\right)\right)
\end{align}
where
$P_{f,i,j}$, $P_{b,i,j}$, and $\Tilde{A}_{i,j}$ are the elements of the normalized forward, backward, and self-adaptive adjacency matrix respectively.

\section{Experiment}

We performed two sets of experiments in this paper.
The first set of experiments was on real-world data to show the superiority of the message-passing flavor for traffic forecasting.
The second set of experiments was on synthetic data to show that the message-passing flavor can capture node interactions better than other flavors.

\subsection{Real-world data}

\subsubsection{Dataset}

We use the most popular dataset from the latest benchmark \cite{jiang2021DL-TrAFF} which also made them publicly available in their GitHub \url{github.com/deepkashiwa20/DL-Traff-Graph}.
\textbf{METR-LA} was collected from loop detectors in LA, USA county highway.
Detailed statistics about the dataset can be found in Table \ref{tab:dataset_rw_desc} in Supplementary Material.

\subsubsection{Experimental Setup}

We used Optuna \cite{optuna_2019} for hyperparameter optimization with a budget of 20 runs for each model.
Following the original setup, we used Adam as the optimizer, and Mean Absolute Error (MAE) as the loss function.
We split the dataset to training/validation/testing set with a ratio of 7:1:2.
We use a standard scaler on the input.




\subsubsection{Results}

\begin{table*}[htb]
\caption{
Performance comparison using real-world data.
The results of our experiments are denoted with asterisk an (*), the rest are baselines taken from a 2021 benchmark \cite{jiang2021DL-TrAFF}.
The 15/30/60 mins column headings refer to the forecasting horizon.
RMSE, MAE, and MAPE are error metrics, lower is better.
The best number in the relevant group is in \textbf{bold}, the second best is \ul {underlined}, and the third best is \textit{italicized}.
%
}
\label{tab:results_rw:la}
\centering
\scriptsize
\begin{tabular}{@{}cl|rrr|rrr|rrr@{}}
\multicolumn{1}{l}{}        &                   & \multicolumn{3}{c}{RMSE}    & \multicolumn{3}{c}{MAE}     & \multicolumn{3}{c}{MAPE (\%)} \\
\multicolumn{1}{l}{} & Model             & 15 mins & 30 mins & 60 mins & 15 mins & 30 mins & 60 mins & 15 mins  & 30 mins  & 60 mins \\ \midrule
\multirow{12}{*}{\rotatebox[origin=c]{90}{METR-LA}}   & HistoricalAverage & 14.737  & 14.737  & 14.736  & 11.013  & 11.010  & 11.005  & 23.34    & 23.34    & 23.33   \\
                            & CopyLastSteps     & 14.215  & 14.214  & 14.214  & 6.799   & 6.799   & 6.798   & 16.73    & 16.73    & 16.72   \\
                            & LSTNet            & 8.067   & 10.181  & 11.890  & 3.914   & 5.129   & 6.335   & 9.27     & 12.22    & 15.38   \\
                            & STGCN             & 7.918   & 9.948   & 11.813  & 3.469   & 4.263   & 5.079   & 8.57     & 10.70    & 13.09   \\
                            & DCRNN             & \textit{7.509}   & 9.543   & 11.854  & 3.261   & 4.021   & 5.080   & 8.00     & 10.12    & 13.08   \\
                            & Graph WaveNet     & 7.512   & \textit{9.445}   & \textit{11.485}  & \textit{3.204}   & \textit{3.922}   & \textit{4.848}   & \textit{7.62}     & \textit{9.52}     & \textit{11.93}   \\
                            & ASTGCN            & 7.977   & 10.042  & 12.092  & 3.624   & 4.514   & 5.776   & 9.13     & 11.57    & 14.85   \\
                            & GMAN              & 8.869   & 9.917   & 11.910  & 4.139   & 4.517   & 5.475   & 10.88    & 11.77    & 14.10   \\
                            & MTGNN             & 7.707   & 9.625   & 11.624  & 3.277   & 3.999   & 4.867   & 8.02     & 10.00    & 12.17   \\
                            & AGCRN             & 7.558   & 9.499   & 11.502  & 3.292   & 4.016   & 4.901   & 8.17     & 10.16    & 12.43   \\
                            & GAT WaveNet*      & \ul{5.303}   & \ul{6.388}   & \ul{7.595}   & \ul{2.759}   & \ul{3.173}   & \ul{3.668}   & \ul{7.14}     & \ul{8.58}     & \ul{10.12}   \\
                            & MP WaveNet*       & \textbf{5.209}   & \textbf{6.247}   & \textbf{7.412}   & \textbf{2.712}   & \textbf{3.093}   & \textbf{3.556}   & \textbf{6.84}     & \textbf{8.23}     & \textbf{10.00}   \\ \bottomrule
\end{tabular}
\end{table*}

The results of our experiments are being compared with the results from a recent benchmark \cite{jiang2021DL-TrAFF} in Table \ref{tab:results_rw:la}. 
Firstly, not including results from our experiments, Graph WaveNet outperforms all the other models more often than not.
In METR-LA, GAT WaveNet significantly outperformed the original Graph WaveNet, and yet still outperformed by MP WaveNet.
We attribute these improvements to GAT capabilities to dynamically adjust the edge weights and MPNN capabilities to capture node interaction.








\subsection{Synthetic data: Root Mean Square of neighbor pairs in a Graph (RMSG)}
\label{sec:RMSG}

To show that the improvements of the MP WaveNet was due to its capacity in capturing complex inter-node interactions,
we tested the different flavors of GNN on a synthetic graph attribute regression task:
Root Mean Square of neighbour pairs in a Graph (RMSG).
The motivation behind this task is to get simple non-linear interactions between a pair of neighbour.

\subsubsection{Task and Dataset}

We constructed a graph with 100 nodes.
10\% of all possible edges are connected; the edges are not directed and are unweighted.
The node feature is a single scalar distributed as follows:
$ x_i = \mathcal{U}_{[-2,2]} $,
where $x_i$ is the attribute of node $i$,
and $\mathcal{U}_{[-2,2]}$ a continuous uniform distribution
with a lower and upper bound of $-2$ and $2$ respectively.

The node label is the root mean square of the product of the features between a node and its neighbors:
$
y_i =
\sqrt{
\frac{1}{n}
\sum_{j \in \mathcal{N} }^{}
(x_i x_j)^2
}
$,
where $y_i$ is the label of node $i$,
and $\mathcal{N}$ is a set containing all the nodes adjacent to node $i$.
The label is designed to have simple interaction, that is multiplications, followed by a simple non-linearity, that is quadratic.
We used the root mean square as the aggregation, as we also use it as the loss function.
This ensures that the behavior that the GNN needs to capture is simply the non-linear interactions, as the aggregation is trivial.

\subsubsection{Experimental Setup}

A new graph is generated for each new data point, with new features and a new adjacency matrix.
The loss function is RMSE.
Each GNN flavor is trained on $1,048,576 = 2^{20}$ datapoints,
validated on $104,857$ datapoints,
and tested on $2^{20}$ datapoints.
The hyperparameters are optimized using Optuna \cite{optuna_2019} with a budget of 20 runs.
The performance is based on the average and standard deviation of five runs.
At every run, a new random seed is used to generate a new graph and initialize the model.
To evaluate the performance, we used three metrics:
Root Mean Square Error (RMSE),
Mean Average Error (MAE),
and Coefficient of Determination ($R^2$).
They are defined in the Supplementary Material \ref{sec:metrics}.

\subsubsection{Models}

There will be four different models compared:
the \textit{average} as the baseline comparison,
as well as the three GNN flavors models.

In the \textit{average} model, the label is the same for the entire graph, and it is the average of the label of the training set:
$ \textbf{h}_n = \frac{1}{N} \sum_{n=1}^{N} \mathbf{y}_n $
As a representative for the
convolutional, attentional, and message-passing flavors, we pick
GCN \cite{kipf2017GCN}, GAT \cite{velickovic2018GAT}, and MPNN \cite{gilmer2017MPNN} models for each flavor respectively.
We pick GCN \cite{kipf2017GCN} because this is the undirected version of diffusion convolution in Graph WaveNet.

To adjust the models for the RMSG task, we make the following changes:
In GCN and GAT, we remove the typical softmax layer at the end because this is a regression task.
We added MLP with one hidden layer as the encoder and decoder.
We also added a self-loop in the adjacency matrix by adding an identity matrix to the adjacency matrix.

\subsubsection{Results}

\begin{table}[htb]
\centering
\caption{
Performance comparison of different models representing different flavors of GNN on the RMSG synthetic dataset based on five runs with different random seeds.
The top number at each cell is the average performance while the bottom number is the standard deviation.}
\label{tab:results_synth}
\begin{tabular}{@{}cccc@{}}
\toprule
Model    & RMSE                                                       & MAE                                                        & $R^2$                                                          \\ \midrule
Average  & \begin{tabular}[c]{@{}c@{}}0.69533\\ $\pm$0.000048\end{tabular} & \begin{tabular}[c]{@{}c@{}}0.59113\\ $\pm$0.000048\end{tabular} & \begin{tabular}[c]{@{}c@{}}-0.00003\\ $\pm$0.000002\end{tabular} \\[2.5ex]
GCN [\cite{kipf2017GCN}] & \begin{tabular}[c]{@{}c@{}}0.6765935\\ $\pm$0.0143293\end{tabular} & \begin{tabular}[c]{@{}c@{}}0.5727497\\ $\pm$0.0103641\end{tabular} & \begin{tabular}[c]{@{}c@{}}0.0527333\\ $\pm$0.0404995\end{tabular} \\[2.5ex]
GAT [\cite{velickovic2018GAT}] & \begin{tabular}[c]{@{}c@{}}0.1685125\\ $\pm$0.1851888\end{tabular} & \begin{tabular}[c]{@{}c@{}}0.1318246\\ $\pm$0.1613489\end{tabular} & \begin{tabular}[c]{@{}c@{}}0.8774309\\ $\pm$0.3083798\end{tabular}  \\[2.5ex]
MPNN [\cite{gilmer2017MPNN}] & \begin{tabular}[c]{@{}c@{}}0.01509\\ $\pm$0.002373\end{tabular} & \begin{tabular}[c]{@{}c@{}}0.00965\\ $\pm$0.001814\end{tabular} & \begin{tabular}[c]{@{}c@{}}0.99952\\ $\pm$0.000147\end{tabular}  \\ \bottomrule
\end{tabular}
\end{table}

The result summary of five runs on each model is shown in Table \ref{tab:results_synth}.
We use \textit{average} model metrics as the baseline.
As expected, the $R^2$ is practically zero.
Note that this number can be negative when the sum of squared error in the numerator of equation \ref{eq:r2} is bigger than the standard deviation of the data in the denominator.
GCN did not manage to learn.
This is shown by the fact that it has similar RMSE and MAE with the \textit{average} model.
It also has an $R^2$ that is very close to zero.
Since past literature \cite{kipf2017GCN} has shown that GCN can deal with non-linearity, this result shows that GCN failed to capture node interaction.

GAT performed better with $R^2$ about 0.88.
It also has a significantly lower RMSE and MAE compared to the GCN and the baseline \textit{average} model.
GAT performance is also less reliable, shown by the high standard deviation across all metrics.
We attributed these improvements due to the attentional flavor capability to use node interactions to dynamically adjust the weight of the contribution from each node.

Only MPNN managed to learn the RMSG task completely RMSE and MAE approximately equal to zero and $R^2$ approximately equal to one.
The result of this set of experiments on the synthetic data shows that there exists a situation involving node interactions where MPNN, representing message-passing flavor, is the only flavor that is capable of fully learning the underlying patterns in the data.



\section{Conclusion}
Among the many factors that influence the future speed in a road network, we argued that node interaction is a plausible factor.
Moreover, among the three different flavors of GNN, we also argued that message-passing is the most appropriate flavor to capture node interaction.
Our experiments on real-world data show the superiority of the message-passing flavor.
We also did additional experiments on the RMSG task to contrast the capabilities of the three flavors of GNN with respect to capturing node interactions, concluding that message-passing is superior, not only in terms of losses and metrics but also in capturing the distribution.

\bibliographystyle{splncs04}
\bibliography{1bib}

\pagebreak
\appendix
\begin{center}
\textbf{\large Supplementary Material}
\end{center}

\section{Traffic Forecasting Visual Problem Definition}

\begin{figure}[htbp]
     \centering
     \begin{subfigure}[b]{\columnwidth}
         \centering
         \includegraphics[width=.6\columnwidth]{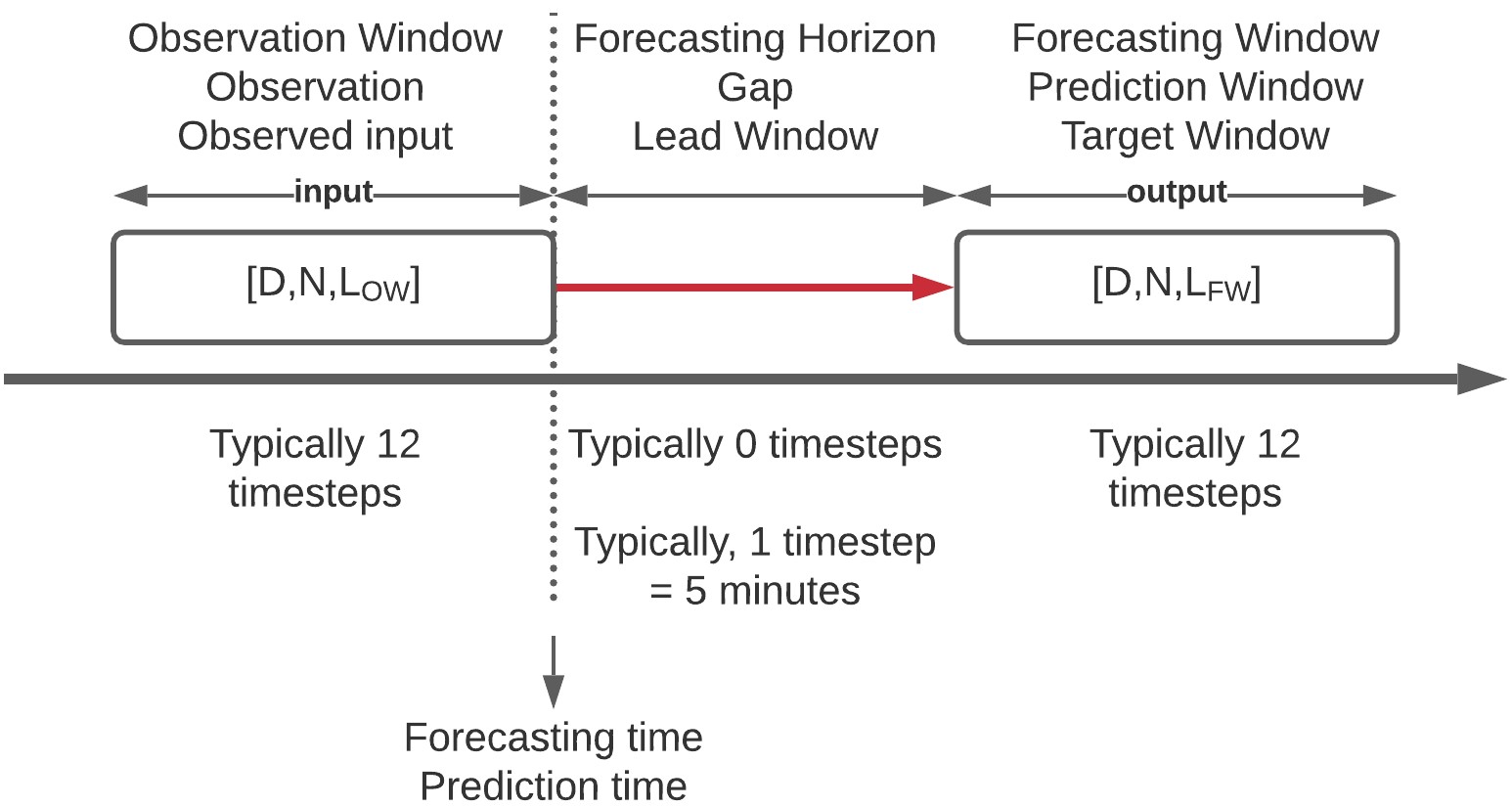}
     \end{subfigure}
     \caption{
     Visual problem definition of traffic forecasting.
     It shows the tensor shape of the input and the output.
     This also shows alternate names of some of the components.
     }
     \label{fig:problem_def}
\end{figure}

The traffic forecasting task is similar to a multivariate time-series forecasting task.
One important difference is that there exist spatial relationships between the different time series.
This is because a single time series in traffic forecasting represents traffic data from a short road segment or a sensor.
A single location can capture more than one traffic metric, such as speed or count.
In the literature, however, most works only use a single metric.
This task can be visualized in Figure \ref{fig:problem_def}.
The formal definition is presented in Section \ref{sec:def}.

\section{Graph WaveNet} \label{apx:gwn}

The overview of Graph WaveNet is shown in Figure \ref{fig:gwn_arch} and the link to the code by the original authors is \url{github.com/nnzhan/Graph-WaveNet}.
The framework consists of an MLP encoder, an alternating temporal and spatial module, and an MLP decoder.
For the temporal module, they used Gated-Temporal Convolutional Network (G-TCN) \cite{van2016wavenet}, and for the spatial module, they used diffusion convolution \cite{li2018DCRNN}.
There are also residual connections that go into the temporal modules, and skip connections from the output of temporal modules.
The skip connections are aggregated with concatenation before going into the final MLP decoder.

A G-TCN module consisted of two parallel 1D convolutions.
The first convolution acts as a filter that projects the input to the output space.
The second convolution acts as a gate that determines
the ratio of information passed to the next layer.
Formally:
$
\mathbf{h}^{l+1} =
\tanh(\mathrm{TCN}(\mathbf{h}^l))+
\sigma(\mathrm{TCN}(\mathbf{h}^l))
$
where
$\mathbf{h}^l$ and $\mathbf{h}^{l+1}$ are the input and output to the $l^\mathrm{th}$ module respectively,
$\sigma(\cdot)$ is the sigmoid function, and
$\mathrm{TCN}(\cdot)$ is a dilated 1D convolution \cite{van2016wavenet}.
Note that the two $\mathrm{TCN}(\cdot)$ have different sets of parameters.

\section{RMSG Metrics definitions}
To evaluate the performances of the RMSG task, we use three metrics:
Root Mean Square Error (RMSE),
Mean Average Error (MAE),
and Coefficient of Determination ($R^2$).
defined as follows:
\begin{align}
    RMSE &= \sqrt{ \frac{1}{N} \sum_{n=1}^{N} (\mathbf{y}_n-\mathbf{h}_n)^2 }, \\
    MAE &= \frac{1}{N} \sum_{n=1}^{N} { \left| \mathbf{y}_n-\mathbf{h_n} \right| }, \\
    R^2 &=  1-\frac
    {\sum_{n=1}^{N} \left( \mathbf{y}_n-\mathbf{h}_n \right) ^2}
    {\sum_{n=1}^{N} \left( \mathbf{y}_n- \left (\sum_{n=1}^{N} \mathbf{y}_n \right) /N \right) ^2}
    \label{eq:r2}
\end{align}
where $\mathbf{y}$ is the actual label,
$\mathbf{h}$ is the predicted value,
and $N$ is the number of datapoints.

\pagebreak
\section{Detailed dataset statistics} \label{sec:metrics}

The detailed statistics of the real-world dataset used in this paper are presented in Table \ref{tab:dataset_rw_desc}.
This data is collected from traffic loop detectors in Los Angles, California, USA.
This data is first collected by \cite{jagadish2014big} and first used for traffic forecasting by \cite{li2018DCRNN}.


\begin{table}[ht]
\caption{Detailed statistics on the real world datasets.}
\label{tab:dataset_rw_desc}
\centering
\setlength{\tabcolsep}{.6em}
\begin{tabular}{@{}cc|r@{}}
\toprule
\textbf{}                 & \textbf{Dataset}     & \multicolumn{1}{c}{\textbf{METR-LA}} \\ \midrule
\multirow{2}{*}{Spatial}  & Nodes                & 207                                  \\
                          & Edges                & 1,515                                \\ \midrule
\multirow{5}{*}{Temporal} & Duration (timesteps) & 34,272                               \\
                          & Duration (days)      & 121                                  \\
                          & Time start           & 01-Mar-12                            \\
                          & Time end             & 30-Jun-12                            \\
                          & Granularity (mins)   & 5                                    \\ \midrule
\multirow{8}{*}{Speed (mph)}    & Min                  & 0.00                           \\
                          & Q1                   & 57.13                                \\
                          & Median               & 63.22                                \\
                          & Mean                 & 58.46                                \\
                          & Q3                   & 66.50                                \\
                          & Max                  & 70.00                                \\
                          & Standard Deviation   & 20.26                                \\
                          & Missing values       & 8.82\%                               \\ \midrule
Size                      & Entry                & 7,094,304                            \\
                          & Compressed (MB)      & 54                                   \\ \bottomrule
\end{tabular}
\end{table}

\pagebreak

\pagebreak
\section{Analysing Model Size on the Synthetic dataset}

\begin{figure}[ht]
    \centering
    \begin{subfigure}[c]{.49\columnwidth}
        \includegraphics[width=\columnwidth]{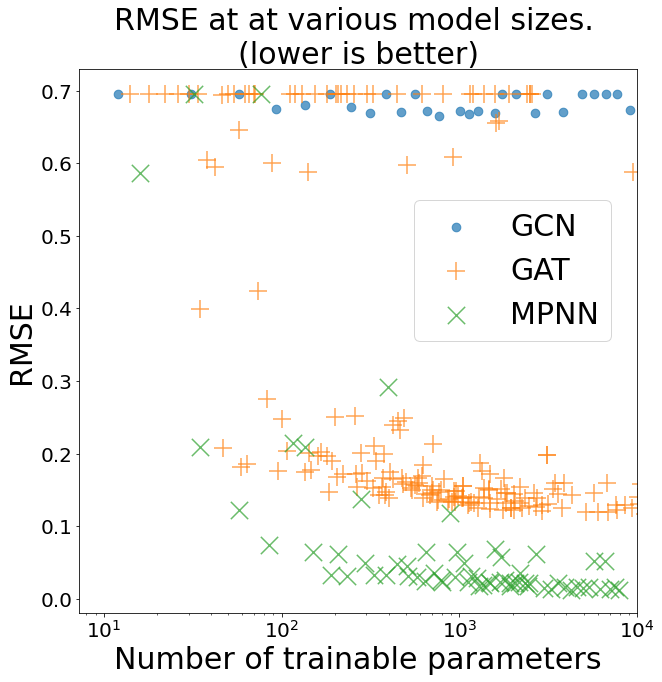}
    \end{subfigure}
    \begin{subfigure}[c]{.49\columnwidth}
        \includegraphics[width=\columnwidth]{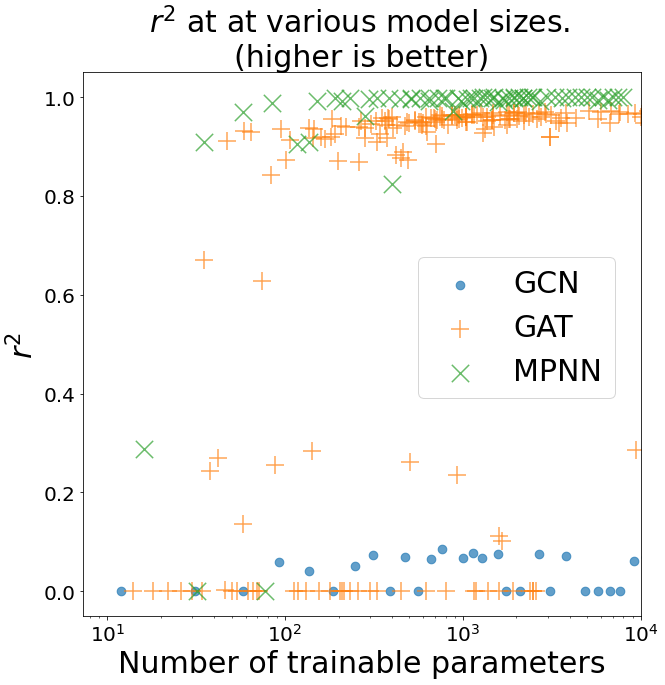}
    \end{subfigure}
    \caption{Performance comparison across different sizes}
    \label{fig:pareto}
\end{figure}

We ran additional experiments to analyze the effect of different model sizes (Figure. \ref{fig:pareto}).
GCN did not manage to learn regardless of the model size.
GAT performances improved with the model size until they plateaued at a sub-optimal level.
GAT was unstable, sometimes failing to learn the behaviors, which is consistent with our finding that it has a high standard deviation in Table \ref{tab:results_synth}.
MPNN performed better across different model sizes and managed to learn the complete behavior with only a few hundred trainable parameters.
At an equal amount of trainable parameters, MPNN always outperformed the others.
It was also more stable than GAT.
These show that the advantage of the message-passing flavor is not simply about more efficiency given the same model size, yet there are certain behaviors, such as where the non-linear interaction between nodes is important, that cannot be learned otherwise.

\pagebreak
\section{Residual Analysis on the Synthetic dataset}

We further analyse the residuals of the three different models.
We define residual as
$
\mathrm{residual}_i = y_i - h_i
$ where
$y_i$ is the label and
$h_i$ is the predicted value.

\begin{figure*}[ht]
    \centering
    \includegraphics[width=\textwidth]{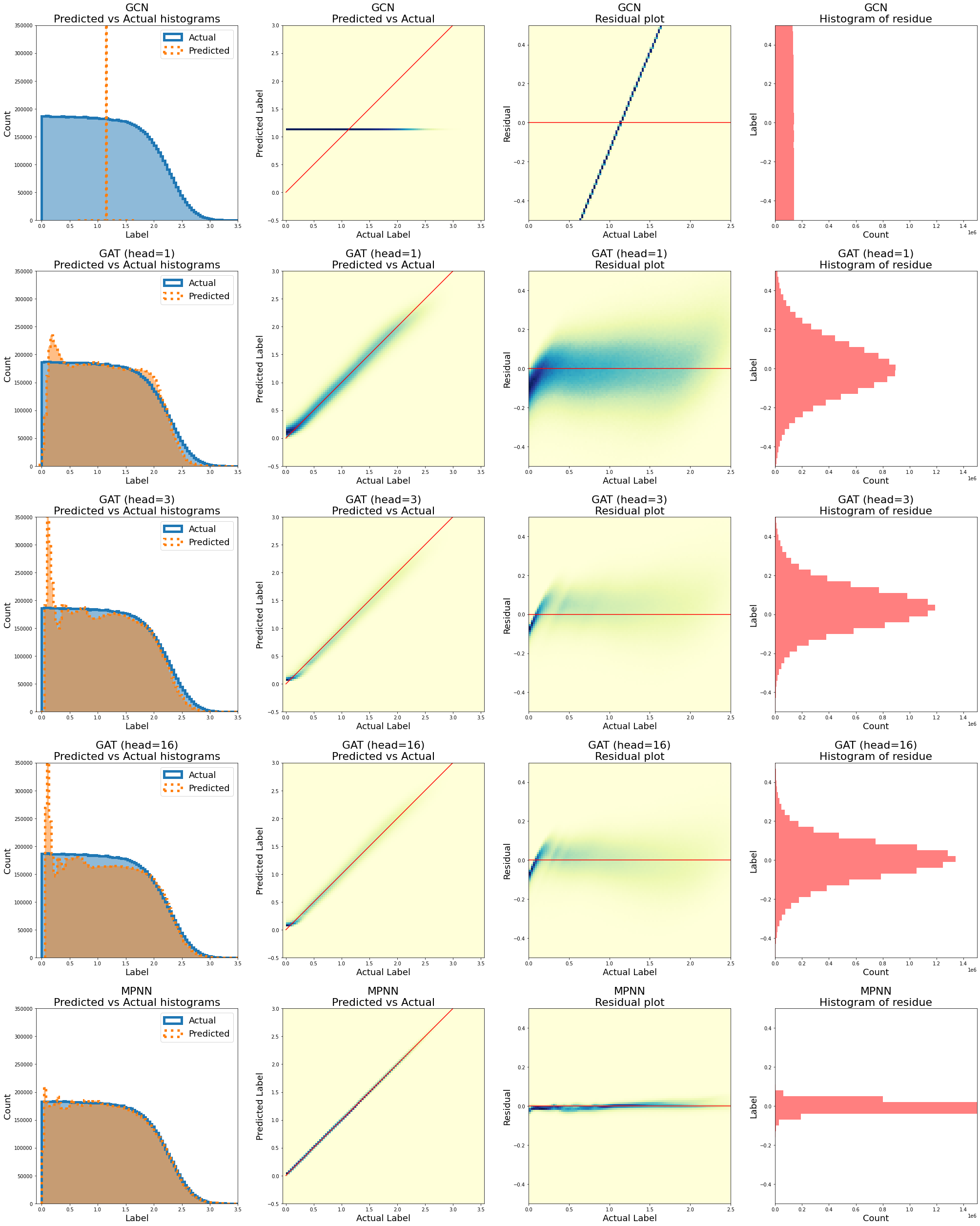}
    \caption{Residual analysis of the different flavors of GNN on the synthetic RMSG task.}
    \label{fig:synth_res}
\end{figure*}

Figure \ref{fig:synth_res} shows the result of the residual analysis.
Each row is a different model, starting from the top: GCN, GAT with 3 different numbers of heads, and MPNN.

The first column compares the distribution of the actual label and the predictions made by the different models.
GCN experienced a mode collapse, predicting similar labels all the time as it does not has node interaction terms at all.
This is also shown in the other columns.
This explains why it has the same performance as the \textit{average} model.

The right-most column shows the distribution of the residuals.
Except for GCN, the distribution is a normal one, as expected from an RMSE loss.
Increasing the number of heads in a GAT decreases the spread of the residuals, as expected.
However, even with 16 heads, it is still not as narrow as MPNN.

The symmetric distribution of the residue hides the systematic error of the GAT models.
The residual plots in the third column show that GAT is systematically over-predicting (negative residual) when the label is small, and systematically under-predicting (positive residuals) when the label is high.
Although the overall error is reduced with more heads, the systematic error persists.
In contrast, MPNN has much smaller systematic errors in the residual plot.

Back to the first column, it shows that GAT has difficulties in approximating the label distribution.
Worse, increasing expressive power from more heads, although reduces the total loss, causes the distribution of the predictions to diverge even further.
In contrast, MPNN predictions distribution is closer to the label distribution.

\end{document}